\documentclass[11pt]{article}
\usepackage{hyperref}

\title{Low-Shot Classification: A Comparison of Classical and Deep Transfer Machine Learning Approaches}
\author{Peter Usherwood \\ \href{mailto:peter.usherwood@kantar.com}{peter.usherwood@kantar.com} 
   \and Steven Smit \\  \href{mailto:steven.smit@kantar.com}{steven.smit@kantar.com} }

\usepackage[margin=2.2cm]{geometry}
\usepackage{ctable}
\usepackage{apacite}
\usepackage{wrapfig}
\usepackage{multirow}
\usepackage{makecell}
\begin{document}
\maketitle
\begin{abstract}
Despite the recent success of deep transfer learning approaches in NLP, there is a lack of quantitative studies demonstrating the gains these models offer in low-shot text classification tasks over existing paradigms. Deep transfer learning approaches such as BERT and ULMFiT demonstrate that they can beat state-of-the-art results on larger datasets, however when one has only $100$-$1000$ labelled examples per class, the choice of approach is less clear, with classical machine learning and deep transfer learning representing valid options. This paper compares the current best transfer learning approach with top classical machine learning approaches on a trinary sentiment classification task to assess the best paradigm. We find that BERT, representing the best of deep transfer learning, is the best performing approach, outperforming top classical machine learning algorithms by $9.7\%$ on average when trained with $100$ examples per class, narrowing to $1.8\%$ at $1000$ labels per class. We also show the robustness of deep transfer learning in moving across domains, where the maximum loss in accuracy is only $0.7\%$ in similar domain tasks and $3.2\%$ cross domain, compared to classical machine learning which loses up to $20.6\%$.
\end{abstract}

\twocolumn
\sloppy
\section{Introduction} \label{introduction}

Transfer learning in the Natural Language Processing (NLP) field has advanced significantly in the last two years, introducing fine-tuning approaches akin to those seen in computer vision some years earlier \cite{imgnet}. This growth originated from feature-based transfer learning, which in the form of word embeddings has been in use for some years, particularly driven by \cite{mikolov2013}. As part of this new wave, we have seen advancements in feature-based transfer learning in the form of ELMo \cite{elmo}. In addition a characteristic trend in this wave of transfer learning models is a class of algorithms that primarily focus on a fine-tuning approach, where a base language model \cite{bengio2003neural} is trained and then fine-tuned on a target task. This base language model is typically very large ($100M+$ parameters) and takes a relatively long time to train. However, the fine-tuning task is usually much quicker to train as only a few parameters are added to the model, typically a single dense layer to the end of a multilayer LSTM or Transformer \cite{transformer}. The model continues training either all, or part, of the network, but this is typically on much less data and for much less time, as only the task specific information is being learned and the general "understanding" of the language is transferred.

These approaches have, on multiple occasions, broken the state-of-the-art records (SOTAs) across the board on a range of NLP tasks and datasets \cite{bert} \cite{ulmfit}. However, all of these datasets are designed for deep learning: they are typically large enough that they warrant the use of deep learning (5000+ examples per class), without the necessity of transfer learning. It is our view that what transfer learning does, in these cases, is push the boundaries of performance. 

The prevalence of deep learning algorithms in surpassing SOTA records suggests quite clearly that, for the datasets assessed, deep learning surpasses the limits of classical machine learning algorithms in NLP tasks.

Low-shot transfer learning is another use-case for transfer learning in NLP, one of particular interest to companies working with real-world data. Low-shot transfer learning (also referred to as "few-shot") is the use of transfer learning in training models where we have little training data available. This is important as many potential real-world applications of machine learning NLP do not have access to sufficiently large datasets to train deep learning algorithms, and obtaining such a dataset can often be too expensive or time consuming.

Howard \& Ruder \citeyear{ulmfit} note, and Devlin et. al. \citeyear{bert} hypothesize that their respective approaches can be used with low quantities of data to give good results. However, in sources such as \cite{ulmfit}, results on low-shot learning are presented relative to training deep models from scratch, but as mentioned in \cite{Goodfellow:2016:DL:3086952}, deep learning generally only achieves reasonable performance at about $5000$ examples per class and is therefore not necessarily the best paradigm at these scales.  This is shown quantitatively in \cite{deepvssvm} where, at scales of $~2000+$ labels per class, an SVM outperforms several deep learning approaches on text classification tasks. As such, we propose that to evaluate the low-shot learning benefits of deep transfer learning models, we should in fact look at performance against the strongest classical machine learning methods. However, we have yet to find a comprehensive quantitative study performing this analysis and show that low-shot transfer learning in NLP is actually the optimal approach when dealing with small quantities of data.

In this paper we attempt to answer this question in the context of classification tasks. What is the best paradigm to use in the case where we have $100-1000$ labelled training examples per class - classical machine learning or deep transfer learning? We seek to compare the best-in-class approaches from both deep transfer learning and classical machine learning by training a variety of models and evaluating by analysing intra-domain and inter-domain performance (details in section \ref{datasets}). 

The choice of $100-1000$ is motivated by the amount of data feasible for companies and researchers to tag in-house, as well as the scale of data occurring organically through other means. For example, in marketing these figures typically represent the base sizes of surveys that can be used as training data.

The rest of this paper is laid out as follows. Section \ref{datasets} details the datasets we use. Section \ref{methodology} looks at the methodology used to evaluate the optimal paradigm. In section \ref{related} we present the algorithms we use to test, along with related work influencing our choices in selecting those models. Section \ref{experiments} details our experiments including choosing the optimal configuration of hyperparameters and preprocessing for each algorithm. In section \ref{results} we present the results followed by our comments and conclusions. Finally, we highlight a few key points and considerations worthy of mention for the two paradigms in \ref{discussion}.

\section{Datasets} \label{datasets}

We have sourced a range of publicly available datasets for classification tasks in an attempt to remove any potentially unknown biases of one particular set. However, to aid in our goal of viewing cross-dataset and cross-domain performance, we have focused on sentiment based classification. This is one of the more popular classification problems as well as being one of critical importance for many companies in areas such as chatbots and social media marketing. Potential biases include the ability for a certain task to be predictable based off of a few low level features (making the task more trivial), or similar data having been used in the pre-training of the deep transfer learning approach, tainting the test set. 

The datasets we consider fall into two domains: Amazon reviews ($A$), and Twitter ($T$). The first category consists of 3 Amazon datasets, one consisting of movie reviews, and two of product reviews from different product categories. Whilst these are very "real world" datasets, we describe this domain as clean data. These datasets typically have similar medium-length documents of $100-300$ words, and are the kinds of datasets typically used in evaluating the performance of deep transfer learning: Pang, Lee \% Vaithyanathan \citeyear{imdbdata} use IMDB and Howard \& Ruder \citeyear{ulmfit} use SST-2 movie reviews.

The second domain has datasets sourced from Twitter, a social data source that differs in a few key properties from the Amazon sets. The vocabulary is much broader given the amount of slang, abbreviations, and the unique way in which hashtags are used grammatically. In addition, Twitter datasets typically will have a stronger prevalence of emoji than in other domains, although in these datasets emoji were already removed. These can make Tweets much harder to classify, particularly for deep transfer learning models that have pre-defined vocabularies. BERT relies on WordPiece embeddings which makes it more robust to new vocabularies \cite{wordpieceembeddings}, although it still can not handle emoji. On the other hand, approaches similar to ULMFiT rely on a set word-token vocabulary defined by the training set used in pre-training, which for ULMFiT is wikitext-103 (a wikipedia based text) by default, so this will struggle both with new vocabulary and emoji. We hypothesise these models will suffer a greater loss in accuracy on these Twitter datasets than the classical algorithms because of this fixed vocabulary limitation.

Below we introduce the five datasets we use. Three from the Amazon reviews domain: Amazon Movie Reviews, Amazon Book Reviews, Amazon Health and Personal Care Product Reviews which we will refer to later in the paper as $A1$, $A2$, and $A3$ respectively. The two datasets from the Twitter domain are both from SemEval 2017, we use subtask \textit{a} and subtask \textit{ce}, which we will refer to as $T1$ and $T2$ respectively.

\subsection{Amazon Movies $A1$} \label{amazonmovies}
The Amazon movies dataset sourced from \cite{amazonmoviesdata} is a huge collection of movie reviews from Amazon, including reviews made up to October 2012. We use a random subset of this for our purposes. All reviews are on a five point scale which we re-sample to a three point scale by binning 2 star with 1 star and 4 star with 5 star reviews. This is a procedure we use throughout this work to align all of our datasets onto a three point sentiment scale of negative, neutral and positive. In this dataset we also have knowledge of which product each review belongs to, so the train, validation, and test sets are split out so that no product appears in two or more sets.

\subsection{Amazon Books $A2$} \label{amazonbooks}
The second dataset we consider is from the Amazon product review database \cite{amazondata}, and contains reviews of books. This dataset was chosen as it is fairly similar to that of the Amazon movies, whilst still being in a different domain. This makes it ideal in helping us avoid biases relating to the specificity of one dataset while training a very similar task. Its similarity also makes it a perfect candidate to test how well classifiers perform cross-domain in a best case scenario. 

The dataset is structured similar to that of the movies, with a star rating from 1 to 5 giving us a five point sentiment scale which we resample to three point. The review text also contains medium-length documents similar to that of $A1$. It also contains information about which product is being reviewed, so to ensure there is no information leakage into the test set, we ensure every book in each set is exclusive.

\subsection{Amazon Health and Personal Care $A3$} \label{amazonhealth}
This dataset is almost equivalent to the above in terms of set-up with the reviews instead focusing on health and beauty products.

\subsection{SemEval 2017 Subtask A $T1$} \label{semevala}
For our Twitter dataset we use SemEval \cite{semeval2017}. SemEval 2017 Task A comes pre-tagged into negative, positive, and neutral classes so no binning is necessary. This is the only dataset we use that does not contain "product" information so we simply randomly divide the Tweets between the train, validation, and test sets.

\subsection{SemEval 2017 Subtask CE $T2$} \label{semevalce}
Again looking at SemEval 2017 we use their subtask CE datasets to produce this set. The data here comes pre-split although as we have a slightly unusual case here wanting only a specific amount of training data and more test data we shuffle everything and re-split. Here we have product information as all tweets are labelled with a topic, so we divide on that variable. Tweets here are also on a five point scale but again we bin to three point by grouping "very negative" and "negative", and "very positive" and "positive".

\section{Methodology} \label{methodology}

To answer the question as to what is the optimal paradigm to use in a low-shot classification task, we will compare the performance of the best in class approach from classical machine learning and deep transfer learning on various low shot datasets.

The metrics we consider will be the accuracy on a held out test set for each model. In addition we shall also test the models' robustness by examining how they perform making predictions across datasets within a domain (intra-domain), and across domains (inter-domain). Here we define a domain as a set of datasets that share similar properties, in this paper we consider two domains: Twitter (T), and Amazon reviews (A). In the academic literature it is rare to see inter-domain accuracy reported except in the case of specifically designed inter-domain algorithms \cite{crossdomainsent}. However, we see this as a common practice in business: train a single classifier and use it inter-domain. As such we feel it should be evaluated since our end use-case is informing the opinion of which approaches to take when building real-world classifiers. Our intention is that the range of datasets will minimise any bias and make the study as relevant to industry use-cases as possible.

We shall also consider several levels of low-shot transfer learning, taking: 100, 300 and 1000 labelled examples per class (henceforth referred to as $t_{100}$, $t_{300}$, and $t_{1000}$ respectively) in an attempt to guard against the potential of missing the point at which deep learning/classical machine learning surpasses the other in the low-shot context.

For every dataset, we set aside controlled test and validation sets. For $A$ datasets we have $5000$ examples per class in the test and validation sets, $3000$ for $T1$ and $1000$ for $T2$.

All fine-tuning and hyperparameter selection is done on the validation set, and all values displayed in this paper in section \ref{experiments} are from the held-out test sets. The test sets comprise products that are independent of the train and validation sets where the definition of a product is specific to each dataset. For example, in the $A1$ dataset, a product is a specific film. We do this to keep the test as fair as possible, and ensure that if classifiers overfit and learn features such as the name of the film as a defining feature, that it is not rewarded in the evaluation.

\subsection{Metrics} \label{intrometrics}

As our end goal is to simply compare classical machine learning to deep transfer learning we aggregate a lot of the metrics when presenting the results as the individual model-dataset combination is not of key interest. Instead we investigate how well the paradigm does when evaluated on a set task of a set domain. We present six metrics to compare performance, each of these were run per model, per tier, totalling 72 results. The exact metrics and how they are defined from the underlying datasets are given below.

\textbf{$A$ Self}: Average of the three different $A$ trained classifiers accuracies as returned on the same datasets held out test set.

\textbf{$A$ Cross $A$}: Average of the three different $A$ trained classifiers accuracies as returned on the other two $A$ datasets corresponding held out test set.

\textbf{$A$ Cross $T$}: Average of the three different $A$ trained classifiers accuracies as returned on the two $T$ datasets held out test sets.

\textbf{$T$ Self}: Average of the two different $T$ trained classifiers accuracies as returned on the same datasets held out test set.

\textbf{$T$ Cross $T$}: Average of the two different $T$ trained classifiers accuracies as returned on the other $T$ datasets corresponding held out test set.

\textbf{$T$ Cross $A$}: Average of the two different $T$ trained classifiers accuracies as returned on the two A datasets corresponding held out test set.

\section{Models and Related Work} \label{related}
We are trying to compare the best in class approach from classical machine learning and fine-tuning transfer learning, as such we have considered popular high performing models used in other well referenced work, as these represent the types of approaches practitioners will look to. The ones considered in this paper are introduced below.

\subsection{Classical Machine Learning} \label{classicalml}
It is well established that there is no single classical machine learning classifier that consistently achieves the best classification performance. For example between the works in \cite{davidson2017automated} \cite{dadvar2014experts} \cite{dinakar2011modeling} they showed that various classical machine learning approaches all slightly out performed each other. This is a long known phenomenon that all of these models have different strengths depending on the specific task and dataset. As such we have considered two of these (Na\"ive Bayes and SVM) to give a fair representation and alleviate the bias of a single classifier.

\subsubsection{Na\"ive Bayes} \label{intronb}
The Na\"ive Bayes classifier is a probabilistic classical machine learning classification algorithm that has a long history of being used in text classification tasks including sentiment analysis. It is well suited to this task given its speed to run and ability to easily incorporate many features which often occurs with classical NLP approaches. 

However as with most algorithms from classical machine learning they are not competitive on large datasets compared to deep learning models as such we struggled to find an undisputed best in class Na\"ive Bayes approach. We decided to try the approaches outlined in \cite{nb} as in the paper the authors clearly show the benefits of each modification they make which we are able to verify for our data in section \ref{experiments}, they also ran their classifier on a very similar dataset to the ones used in this paper (movie sentiment classification).

\subsubsection{SVM} \label{introsvm}
Support Vector Machines (SVMs) attempt to separate the data by finding the hyperplane in n-dimensional space that maximizes the distance between the closest (support) vectors in the dataset. SVMs also have long been used in text classification problems given they perform well when our number of features is large compared to the number of training examples as is the case in our classical paradigm.

Indeed we see in \cite{deepvssvm} the authors compare an approach using SVM with an n-gram approach to word vector deep learning approaches on datasets that are larger than the smallest dataset considered in this work, and in all balanced cases the SVM n-gram models outperform the deep learning approaches. 

As such we consider the n-gram approach with SVMs however given the results gained in \cite{nb} we also try all of the same additions to the architecture for SVMs that we trial for Na\"ive Bayes to further improve performance, the results of this analysis are shown in section \ref{experiments}.

\subsection{Transfer Learning} \label{transferml}
In the deep transfer learning paradigm we have seen a quick succession of models, as discussion in section \ref{introduction}, released in the last two years each of which has surpassed the previous in terms of performance on deep learning tasks. Recently, the leading approach has been BERT \cite{bert} and hence this will be the principle approach we consider here. Additionally we also look at ULMFiT \cite{ulmfit}, as this was one of the first major breakthrough approaches in NLP transfer learning and is widely used through their dedicated pytorch-based library. We also look to quantify the claims in Howard \& Ruder \citeyear{ulmfit} that it performs well on low-shot tasks. At the time of writing, it appears that Yang et. al. \citeyear{xlnet} is the current SOTA, surpassing BERT. However, this is very recent and we leave evaluation of this as a future area of research to build on the work documented here. Indeed this raises a key point: transfer learning approaches are still not fully developed and with time we expect these approaches to further improve relative to classical machine learning. 

\subsubsection{ULMFiT} \label{introulm}
ULMFiT was introduced by Howard \& Ruder \citeyear{ulmfit} and was one of the first popular applications of fine-tuning transfer learning in NLP. They achieved SOTAs on various classification datasets: AG, DBpedia, Yelp-bi, and Yelp-full. 

Their approach is to use an AWD-LSTM, an architecture originating in the work of \cite{merity2017} as the language model at the core of the model. The model is then trained in 3 stages, first (Stage 1) the core language model is trained on a large general purpose corpus. This is the pre-training step which is ideally done only once per language. Stage 2 is the fine-tuning step where the same core language model continues to be trained but now on the target dataset, this aids the model in learning the nuances of the target task language which ultimately improves results on the final classification. The training of the classification task is the third stage, in which two dense layers are appended to the final hidden layer of the language model and the whole model is trained on the supervised classification task. Advanced techniques such as slanted triangular learning rates and gradual unfreezing are used to negate the problem of catastrophic forgetting, enabling the model to better retain earlier learned information.

Although the model is pre-training dataset agnostic, the current published model was built on wikitext-103. The potential issue with this is choice of pre-training dataset is that it is not very general purpose for many real world applications, such as social datasets. The model is also built using a fixed vocabulary which further limits its generalizability. 

We use this pre-trained AWD-LSTM based model and follow the recommended fine-tuning stages in this paper, whilst verify the choice of all hyperparameters on held out validation sets in section \ref{experiments}.

\subsubsection{BERT} \label{introbert}
BERT (Bidirectional Encoder Representations from Transformers) was introduced in late 2018 by Devlin et. al. \citeyear{bert}. We chose this algorithm for its performance (breaking various SOTAs on NLP tasks, both in classification and other challenges such as question answering) and popularity. Conceptually it is similar to that of ULMFiT, a core language model trained on a large general purpose corpus followed by a stage of task specific fine-tuning to learn a supervised task. BERT has the generalizability to work with classification or sequence based target tasks which gives it further utility, however in this paper we focus on its ability in classification tasks. One important distinction between the two approaches for our purposes is that feature representation in BERT is based on word pieces \cite{wordpieceembeddings} which may afford the model better generalizability than ULMFiT. 

On release BERT was published with two models: BERT-base which uses $12$ transformer layers and has $110M$ parameters, and BERT-large with $24$ layers and $340M$ parameters. In this paper we use BERT-base and follow the suggestions of fine-tuning as given in the original paper. We verify our model in section \ref{experiments} on our validation sets.

The second and final stage of training BERT for classification tasks is to append a single dense layer to the final hidden layer of the language model and continue training.

\section{Experiments} \label{experiments}
We initially setup the models based on the referenced papers and fine-tuning on the validation sets. Given the volume of models and datasets in this work it is unfeasible to fine-tune the hyperparameters for every model to be trained. Instead we originally intended to optimize only one set of hyperparameters and pre-processing stages for each model, and then train all models on this configuration. However when we looked at the $A1$ and $T1$: 100 and 1000 to optimize the hyperparameters we saw significant deviation in the optimal configuration across these sets for the classical machine learning approaches, particularly on the SVM. As such we proceeded having up to four configurations per approach: A $t_{100}$-$t_{300}$, A $t_{1000}$, T $t_{100}$-$t_{300}$, T $t_{1000}$. 

\subsection{Na\"ive Bayes Hyperparameter Search} \label{hypernb}
We mainly followed the approach laid out in Narayanan et. al \citeyear{nb} for our Na\"ive Bayes approach, where negation was handled by merging any tokens that were preceded by any in a set of pre-defined negation terms, with "not\_" as a prefix and removing the original negation term. We also followed Narayanan et. al's \citeyear{nb} suggestion of using a Bernoulli term frequency matrix and including bi-grams and tri-grams, as we found all of these methods independently boosted the accuracy on the validation sets for $A1$. However on $T1$ none of the approaches led to any increase in performance. We also experimented with reducing the number of features $n_{features}=f \cdot tier$ as a hyperparameter, where the grid search begins at the maximum number of features and $f$ is tuned to reduce the number of features downwards, selecting for a more parsimonious model. We present our search on the validation sets in table \ref{tab:2}, also included were the papers original gains on binary sentiment classification. We also tried other approaches such as stemming and feature selection with Part of Speech (PoS) tagging although these showed no benefit on any dataset, although perhaps with more manual feature selections gains could be made.

\begin{table*}[ht!]
\footnotesize
\centering
\begin{tabular}{lccccc}  \specialrule{.1em}{.05em}{.05em}
                          & \multicolumn{1}{l}{$A1$-100} & \multicolumn{1}{l}{$A1$-1000} & \multicolumn{1}{l}{$T1$-100} & \multicolumn{1}{l}{$T1$-1000} & \multicolumn{1}{l}{IMDb \cite{imdbdata} \% binary gain}  \\  \specialrule{.1em}{.05em}{.05em}
Original                  & 48.06                                & 55.73                                 & \textbf{49.35}                            & \textbf{57.64}                             & 23.77                                        \\
Negation                  & 48.54                                & 56.52                                 & 48.29                            & 57.29                             & 9.03                                         \\
Bernoulli                 & 49.26                                & 56.31                                 & 48.77                            & 57.03                             & \multicolumn{1}{l}{-}                        \\
N-grams Bernoulli         & 50.09                                & \textbf{61.20}                                  & 47.85                            & 56.65                             & \multicolumn{1}{l}{-}                        \\
Bernoulli Negated         & 49.97                                & 56.99                                 & 48.23                            & 56.49                             & 0.86                                         \\
N-grams Bernoulli Negated & \textbf{50.19}                                & 61.15                                 & 47.02                            & 56.24                             & 1.54                                        
\\ \specialrule{.1em}{.05em}{.05em}
\end{tabular}
\caption{Hyperparameter search of the best architecture to use for the Na\"ive Bayes approach, with the four models chosen shown in bold.}
\label{tab:2}
\end{table*}

\subsection{SVM Hyperparameter Search} \label{hypersvm}
Looking at SVMs we try to apply a similar approach as we do for Na\"ive Bayes. Here we add the extra hyperparameters of C (the SVM regularization hyperparameter), and the kernel type. We initially select values of $f=max$ and $C=1$ which showed good performance to trial the various additions to the architecture. Once we evaluated the best architecture  we then fine-tuned $C$ and $f$. Finally we verified all of the architectures again to check there were no changes in top performance. The final results chosen were $f=max,50,10,10$, $C=1,1,.5,1$, and linear kernels on $A1$-$t_{100}$, $A1$-$t_{1000}$, $T1$-$t_{100}$, $T1$-$t_{1000}$ respectively as shown in table \ref{tab:3}.

\begin{table*}[ht!]
\footnotesize
\centering
\begin{tabular}{lccccc}  \specialrule{.1em}{.05em}{.05em}
                          & \multicolumn{1}{l}{$A1$-100} & \multicolumn{1}{l}{$A1$-1000} & \multicolumn{1}{l}{$T1$-100} & \multicolumn{1}{l}{$T1$-1000}   \\  \specialrule{.1em}{.05em}{.05em}
Original                           & 46.63                                   & 54.76                                    & 44.63                                     & 52.82                                       \\
Negation                           & 46.48                                   & 55.03                                    & 43.67                                     & 52.98                                       \\
Bernoulli                          & 47.51                                   & 55.08                                    & 45.71                                     & 52.82                                       \\
N-grams Bernoulli                  & \textbf{50.30}                                   & 58.10                                    & 43.13                                     & 52.82                                       \\
Bernoulli Negation                 & 47.47                                   & 56.01                                    & 45.04                                     & 52.76                                       \\
N-grams Bernoulli Negation         & 49.59                                   & 58.62                                    & 42.23                                     & 51.93                                       \\
N-grams Bernoulli Negation Stopped & 48.67                                   & 57.80                                    & 43.96                                     & 51.39                                       \\
\multicolumn{1}{l}{Tf-idf}          & 48.88                                   & 59.84                                    & \textbf{48.84}                                     & \textbf{56.97}                                       \\
N-grams Tf-idf                     & 44.23                                   & 46.66                                    & 45.01                                     & 43.86                                       \\
Tf-idf Negation                     & 48.80                                   & 60.13                                    & 48.39                                     & 56.56                                       \\
N-grams Tf-idf Negation             & 47.92                                   & \textbf{61.41}                                    & 46.76                                     & 55.69                                       \\
N-grams Tf-idf Negation Stopped     & 46.45                                   & 60.14                                    & 47.72                                     & 55.41        \\          \specialrule{.1em}{.05em}{.05em}                     
\end{tabular}
\caption{Hyperparameter search of the best architecture to use for the SVM approach, with the four models chosen shown in bold.}
\label{tab:3}
\end{table*}

\subsection{ULMFiT Hyperparameter Search} \label{hyperulm}
One of the immediate benefits of fine-tuned transfer learning approaches is that complex features representations are transferred, this means that a lot of the effort in preprocessing and feature selecting is not required (or even possible) in these cases. Furthermore an additional benefit is that the models are designed to need minimal task specific architecture adjustments again reducing the amount of parameter selection needed on a validation set. In our experiments we use the pre-trained AWD-LSTM weights published with the original paper, and continue with the proposed methodological approach for the classification task, as implemented in the \textit{fastai} python package (Fast.ai are behind the ULMFiT methodology). 

This leaves us with six choices of hyperparameters: the number of epochs for fine-tuning and classification training, and the corresponding base learning rates and dropout scaling rate. By dropout scaling rates, we mean a scale applied to the packages preset dropout rates. We present the results found on our four searches in the validation sets in table \ref{tab:4}. 

\begin{table}[ht!]
\footnotesize
\centering
        \begin{tabular}{lllll}
        \specialrule{.1em}{.05em}{.05em}
        Dataset    & Epochs & \thead{Learning \\Rate} & \thead{Dropout \\Scaling} & Loss   \\
        \specialrule{.1em}{.05em}{.05em} 
        \multicolumn{5}{c}{Stage 2}                                    \\
        \specialrule{.1em}{.05em}{.05em} \\
        $A1$-$t_100$  & 13     & 0.0025        & 0.5             & 3.55 \\
        $A1$-$t_1000$ & 29     & 0.0003        & 0.5             & 3.79 \\
        $T1$-$t_100$  & 115    & 0.001         & 1               & 3.01  \\
        $T1$-$t_1000$ & 40     & 0.001         & 1               & 3.57 \\
        \specialrule{.1em}{.05em}{.05em}
        \multicolumn{5}{c}{Stage 3}                                    \\
        \specialrule{.1em}{.05em}{.05em} \\
        $A1$-$t_100$  & 12     & 0.0015        & 0.5             & 0.99 \\
        $A1$-$t_1000$ & 23     & 0.0003        & 0.6             & 0.86 \\
        $T1$-$t_100$  & 5      & 0.0015        & 0               & 1.02 \\
        $T1$-$t_1000$ & 17     & 0.0005        & 0.5             & 0.86 \\
        \specialrule{.1em}{.05em}{.05em}
        \end{tabular}
        \caption{Results of the hyperparameter search for Stage 2 \& 3 of the ULMFiT approach.}
\label{tab:4}
\end{table}[t]

It should be noted that in stage two of training a ULMFiT we can train on more domain data than we have labelled. We chose not to do this to give the worse case scenario for practitioners using these models. 

\subsection{BERT Hyperparameter Search} \label{hyperbert}
As noted, one of the benefits of transfer learning is that minimal task specific hyperparameters are needed and we do not need to select features. As such we go with the standard approach used in the original paper, of attaching a single dense layer to the end of the BERT model. We are using the pre-trained BERT-base model: in the original paper, significantly better results were obtained with the larger model.  Due to the use of consumer-grade hardware for this investigation, we will be making use of BERT-base (which already requires 12GB of VRAM). It is suggested that any results here would only be improved upon by practitioners capable of running BERT-large. We use the uncased version.

Similar to ULMFiT, we are left only choosing the learning rate and number of epochs for the classification phase, however in BERT only one phase is run, so we only need to select two hyperparameters. 

In the original BERT paper the authors comment that often hyperparameter tuning on BERT is unnecessary, particularly for larger datasets, and that an ideal search area is with 3 epochs and learning rates between $5 \cdot 10^{-5}$ and $1 \cdot 10^{-5}$. We followed this recommendation testing learning rates of $1 \cdot 10^{-4}, 5 \cdot 10^{-5}, 3 \cdot 10^{-5}, 3 \cdot 10^{-5}$, and $1 \cdot 10^{-5}$ and $3, 4, 5$, and $10$ epochs and found largely the same results. Our best results are shown in table \ref{tab:5}. However as even our bigger datasets are still low-shot and as such this step was still necessary as in the given range we can still see gains of $3\%+$ accuracy.

\begin{table}
\footnotesize
\centering
\begin{tabular}{lccc} \specialrule{.1em}{.05em}{.05em}  
Dataset & \multicolumn{1}{l}{Epochs} & Learning Rate    & \multicolumn{1}{l}{Accuracy}  \\ \specialrule{.1em}{.05em}{.05em}  
$A1$-$t_{100}$  & 4                                    & $2 \cdot 10^{-5}$ & 59.04                        \\
$A1$-$t_{1000}$ & 3                                    & $2 \cdot 10^{-5}$ & 64.95                        \\
$T1$-$t_{100}$  & 5                                    & $2 \cdot 10^{-5}$ & 66.57                        \\
$T1$-$t_{1000}$ & 3                                    & $1 \cdot 10^{-5}$   & 70.33                       
\\ \specialrule{.1em}{.05em}{.05em}  
\end{tabular}
\caption{Final results for the best hyperparameters in BERT after running a grid search across the number of epochs and learning rates.}
\label{tab:5}
\end{table}

\section{Results} \label{results}

\begin{table*}[ht!]
\footnotesize
\centering
\begin{tabular}{lcccccc} \specialrule{.1em}{.05em}{.05em} 
                           & $A$ Self & $A$ Cross $A$ & $A$ Cross $T$ & $T$ Self & $T$ Cross $T$ & $T$ Cross $A$  \\ \specialrule{.1em}{.05em}{.05em} 
ULMFiT & 45.0	& 41.6		&	37.9	& 41.7		& 	41.2	& 36.6      \\
BERT   & 59.5       & 58.8      & 52.3      & 55.5       & 55.4      & 63.4       \\
Na\"ive Bayes                & 49.1       & 45.6      & 36.6      & 46.6       & 46.0      & 37.7       \\
SVM    & 48.8	& 45.8 &	36.2 &	46.4 &	44.6 &	36.3        \\ \specialrule{.1em}{.05em}{.05em} 
Transfer Best           & \textbf{59.5}       & \textbf{58.8}      & \textbf{52.3}      & \textbf{55.5}       & \textbf{55.4}      & \textbf{63.4}        \\
Classic Best          & 49.1       & 45.8      & 36.6      & 46.6       & 46.0      & 37.7        \\ \specialrule{.1em}{.05em}{.05em} 
\end{tabular}
\caption{Final results for $t_{100}$.}
\label{tab:6}
\end{table*}

\begin{table*}[ht!]
\footnotesize
\centering
\begin{tabular}{lcccccc} \specialrule{.1em}{.05em}{.05em} 
                           & $A$ Self & $A$ Cross $A$ & $A$ Cross $T$ & $T$ Self & $T$ Cross $T$ & $T$ Cross $A$  \\ \specialrule{.1em}{.05em}{.05em} 
ULMFiT & 51.0	& 48.2	&	41.2	&	44.1	&	44.5	& 37.7        \\
BERT   & 62.1       & 61.9      & 55.0      & 55.7       & 55.6      & 63.3       \\
Na\"ive Bayes                & 55.8       & 50.0      & 37.3      & 49.5       & 50.1      & 38.7       \\
SVM    & 53.4 &	49.0 &	36.4 &	48.7 &	48.8 &	37.8       \\ \specialrule{.1em}{.05em}{.05em} 
Transfer Best           & \textbf{62.1}       & \textbf{61.9}      & \textbf{55.0}      & \textbf{55.7}       & \textbf{55.6}      & \textbf{63.3}       \\
Classic Best          & 55.8       & 50.0      & 37.3      & 49.5       & 50.1      & 38.7       \\ \specialrule{.1em}{.05em}{.05em} 
\end{tabular}
\caption{Final results for $t_{300}$.}
\label{tab:7}
\end{table*}

\begin{table*}[ht!]
\footnotesize
\centering
\begin{tabular}{lcccccc} \specialrule{.1em}{.05em}{.05em} 
                           & $A$ Self & $A$ Cross $A$ & $A$ Cross $T$ & $T$ Self & $T$ Cross $T$ & $T$ Cross $A$  \\ \specialrule{.1em}{.05em}{.05em} 
ULMFiT & 56.6	& 52.0	& 42.2	& 51.3	& 51.5	& 39.0        \\
BERT   & 63.3	& 63.3	& 55.6	& 55.7	& 55.8	& 63.3      \\
Na\"ive Bayes                & 61.3       & 54.6      & 38.3      & 54.0       & 56.4      & 40.0       \\
SVM    & 61.4 &	54.9 &	39.2 &	52.6 &	54.9 &	40.3       \\ \specialrule{.1em}{.05em}{.05em} 
Transfer Best           & \textbf{63.3}	& \textbf{63.3}	& \textbf{55.6}	& \textbf{55.7}	& 55.8	& \textbf{63.3}       \\
Classic Best              & 61.4       & 54.9      & 39.2      & 54.0       & \textbf{56.4}      & 40.3       \\ \specialrule{.1em}{.05em}{.05em} 
\end{tabular}
\caption{Final results for $t_{1000}$.}
\label{tab:8}
\end{table*}

Referring to the results in tables \ref{tab:6}, \ref{tab:7} \& \ref{tab:8}, there are a few immediately notable results. As mentioned there is no consistently best performing classical machine learning algorithm, in this case we see that Na\"ive Bayes outperforms SVM by a small margin on most metrics and as such in this study it represents the best in class metrics for classical machine learning in most cases. It's margin over SVM is typically small and an expected discrepancy.

In the case of deep transfer learning we can see a huge difference in the performance of BERT and ULMFiT, our first major conclusion is that ULMFiT does not perform adequately at these scales, performing $14.02 \pm 6.18 \%$ worse than BERT, and worse than both classical approaches. As such all of our results and conclusions take the results of BERT as representing the best in class approach from deep transfer learning.

An interesting result is the finding that deep transfer learning is clearly the strongest performing paradigm for $A$ self \& $T$ self at low scales ($t_{100}$). It gains $+10.4\%$ accuracy on $A$ datasets and $+8.9\%$ on $T$ datasets. However, this gain tapers off as one moves to higher training set sizes, and at the $t_{1000}$ level, this gain has fallen away to $+1.9\%$ and $1.7\%$ on $A$ and $T$ respectively. We would expect this trend to swing back in favour of deep transfer learning at a certain point, as the number of examples continues to climb, given that classical machine learning performance is known to plateau with increased data sizes. 

In the authors' opinion, perhaps the key finding here is how robust the transfer learning performance is inter-domain. We can see that at all tiers, transfer learning exhibits practically no loss in performance (max $-0.7\%$) for intra-domain tasks ($A$ cross $A$ and $T$ cross $T$). On the other hand, classical machine learning loses significant performance in $A$ cross $A$ (max $-16.5\%$), although it should be noted that classical machine learning does not generally lose performance in $T$ cross $T$. 

This robustness is further exemplified when considering inter-domain performance. Comparing $A$ cross $T$ to $T$ self and $T$ cross $A$ to $A$ self, the biggest loss in performance the best transfer learning approach is $3.2\%$ on $t_{100}$ and on higher tiers this drops to $0.1\%$, compared to the best classical approach which drops on average $11.4\%$ for $t_{100}$ and $21.1\%$ at $t_{1000}$, not managing to show much improvement over baseline accuracy. 

These results suggest that the deep transfer learning approaches are capable of much deeper and more complex representations, such that they can utilize previously learned features for newer documents, even when the type of document differs significantly in key properties such as length and vocabulary. On the other hand, the classical machine learning approaches are able to perform reasonably well on the target task, but perhaps by learning more superficial patterns that don't transfer to different domains well. This result is particularly useful in industry where one classifier is used on a range of domains (sentiment classifiers being a noteworthy example). When doing so using classical approaches, one must be very careful and understand exactly how, and on what dataset, the accuracy of the classifier was assessed. With transfer learning, this inter-domain usage appears to be a much safer practice. 

We do notice that across the tiers for $T$ self we see a slightly better relative performance for classical machine learning than in the case of $A$, even though it still generally performs worse than transfer learning. As mentioned in section \ref{datasets}, our main hypothesis for this is the unique language (be that slang, misspellings, niche topics) used in $T$ causes the predetermined vocabularies of the language models to miss a lot of the nuance in the text, to varying degrees (BERTs Wordpiece embeddings should make it slightly more adaptable and we see that in the results). We would expect further relative losses in performance should the target domain have even more out of vocabulary tokens such as emoji on these base models. However this potential limitation could be overcome with a different tokenization method and more diverse datasets during the pre-training stage for deep transfer learning.

\section{Discussion} \label{discussion}
Based on the conclusion in section \ref{results}, we would suggest that deep transfer learning is generally the better paradigm in low-shot classification tasks. It is worth noting however that the accuracies achieved are still on the low-end, and may not be at a high enough quality for some applications. 

A core focus in this research is to aid practitioners in choosing which paradigm to use when trying to solve real-world problems, as such particular consideration needs to be paid to the availability of a core language model for the language of the task in question. The original authors of both the ULMFiT and BERT papers only made English models available upon release. Since then, BERT has released a multi-language model, however no dedicated single language core models exist for either BERT or ULMFiT outside of English. This can be a problem as training the core models is time-consuming and expensive, and may not be feasible in many companies. As a corollary point, the base language model must have been pre-trained on a suitably wide dataset, or one that is similar to the target task's, as the vocabulary limitation mentioned in section \ref{results} could play a significant role. This could mean even publicly available base models in the required language are unsuitable if the target task is significantly dissimilar. 

Production resource requirements are another area of concern. The BERT-base model requires 12GB of VRAM to fine-tune and run - at the time of writing this is considered high-end consumer hardware. Other models such as BERT-large or OpenAI GPT-2 \cite{openaigpt2}, are much too big to fit on consumer hardware. This makes it out of reach for many researchers and undesirable for companies who are trying to minimize costs and are often reluctant to build pipelines that rely on expensive GPU compute instances. Contrary to this, classical ML models are typically small and can be trained and run on any modern laptop.

Despite these issues, deep transfer learning does offer cost saving benefits. It saves a lot of human time in two key areas: feature creation and hyperparameter selection. When considering the scalability in application of models throughout a company, these two benefits coupled with the robustness of using models in cross domain applications, can represent much bigger gains in return on investment than what is spent on hardware costs. Furthermore, the hidden representations of deep transfer learning models can also be used in sequence to sequence tasks, further broadening their applicability to a range of tasks faced in practice.

In conclusion, the evidence presented in this work suggests that deep transfer learning is the best approach to use in low-shot learning tasks, owing to the ability to effectively transfer models into different domains, the quick and easy implementation of these freely available models, as well as no longer needing to conduct expensive and time-consuming hyperparameter searches and training schedules. However, the performance achieved on low-shot tasks suggests there is still much work to be done in obtaining high quality classifiers for low-shot task settings. 

\bibliography{references}
\end{document}